\documentclass[10pt,twocolumn,letterpaper]{article}

\usepackage{cvpr}
\usepackage{times}
\usepackage{epsfig}
\usepackage{graphicx}
\usepackage{amsmath}
\usepackage{amssymb}
\usepackage{adjustbox}
\usepackage{pifont}

\newcommand{\tabincell}[2]{\begin{tabular}{@{}#1@{}}#2\end{tabular}}
\newcommand{\querypoint}{{\color{red}\ding{73}}}

\usepackage[breaklinks=true,bookmarks=false]{hyperref}

\cvprfinalcopy 

\begin{document}

\title{PCT: Point Cloud Transformer}

\author{Meng-Hao Guo\\
Tsinghua University\\
{\tt\small gmh20@mails.tsinghua.edu.cn}
\and
Jun-Xiong Cai\\
Tsinghua University\\
{\tt\small caijunxiong000@163.com}
\and
Zheng-Ning Liu\\
Tsinghua University\\
{\tt\small lzhengning@gmail.com}
\and
Tai-Jiang Mu\\
Tsinghua University\\
{\tt\small taijiang@tsinghua.edu.cn}
\and
Ralph R. Martin\\
Cardiff University\\
{\tt\small ralph@cs.cf.ac.uk}
\and
Shi-Min Hu\\
Tsinghua University\\
{\tt\small shimin@tsinghua.edu.cn}
}

\newcommand{\rrm}[1]{{}}
\newcommand{\cjx}[1]{{#1}}
\newcommand{\cjxc}[1]{{}}
\newcommand{\mtj}[1]{{#1}}
\newcommand{\tjc}[1]{{}}
\newcommand{\gmh}[1]{{#1}}
\newcommand{\ck}[1]{{ #1}}
\newcommand{\lzn}[1]{{#1}}


\maketitle


\begin{abstract}

The irregular domain and lack of ordering {make it challenging} {to design deep neural networks} for   point cloud processing.
    This paper presents a {novel} framework named \emph{Point Cloud Transformer(PCT)} for point cloud learning. PCT is based on Transformer, which achieves huge success in natural language processing and displays great potential in image processing.
    It is inherently permutation invariant for processing a sequence of points, making it well-suited for point cloud learning.
    {To better capture local context within the point cloud, we enhance input embedding with the support of farthest point sampling and nearest neighbor search.}
    Extensive experiments demonstrate that the PCT achieves the state-of-the-art performance on shape classification, part segmentation, semantic segmentation and normal estimation tasks.
\end{abstract}

\section{Introduction}

 {Extracting semantics directly from a point cloud is an urgent requirement in some applications such as robotics, autonomous driving, augmented reality, etc.}
 {Unlike 2D images,} point clouds are disordered and unstructured, {\mtj{making} it challenging to design neural networks to process them.}
Qi et al.~\cite{qi2016pointnet} pioneered PointNet for {feature learning on} point clouds {by using multi-layer perceptrons (MLPs), max-pooling and rigid transformations to ensure invariance under permutations and rotation.}
    {Inspired by} strong progress {made by convolutional neural networks (CNNs)} in the field of {image processing},  many {recent} {works}~\cite{Tchapmi2017segcloud,li2018pointcnn,Atzmon2018point,Wu2019pointconv} have considered
to {define} convolution operators {that can aggregate local features} for point clouds.
{These methods either reorder the input point sequence or voxelize the point cloud}
to obtain a canonical domain for convolutions.

\begin{figure}[t]
    \centering
    \includegraphics[width=\columnwidth]{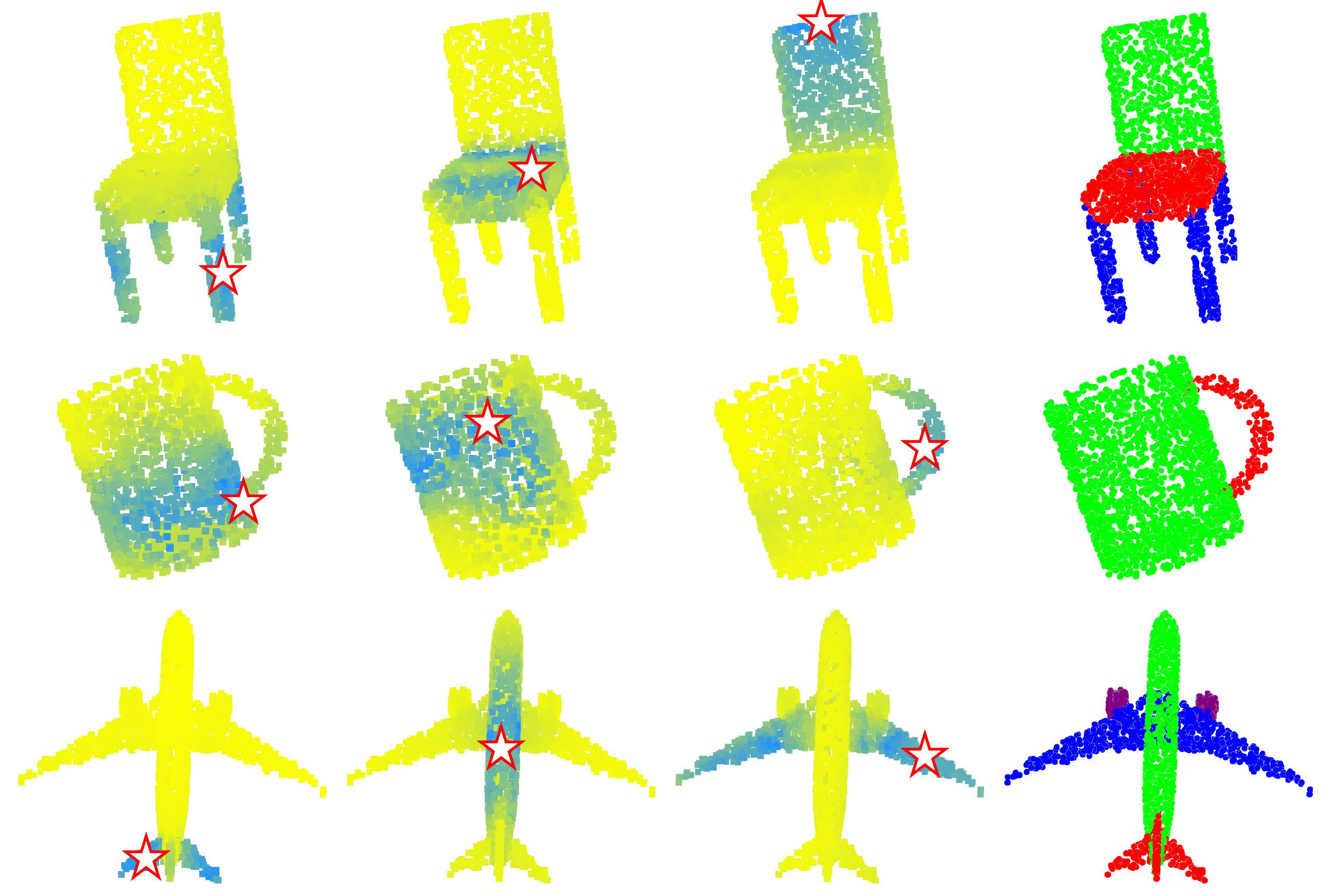}
    \caption{
        Attention map and part segmentation generated by PCT.
        First three columns: point-wise attention map for different query points (indicated by \querypoint), yellow to blue indicating increasing attention weight.
        Last column: part segmentation results.
    }
    \label{fig:attention}
    \vspace{-3ex}
\end{figure}


Recently, \textit{Transformer}~\cite{Vaswani2017attention}, the dominant framework in natural language processing, has been applied to image vision tasks, giving better performance than \gmh{popular} convolutional \gmh{neural networks}~\cite{Dosovitskiy2020ViT,Wu2020visual}.
Transformer is a decoder-encoder structure that contains three main modules for input (word) embedding, positional (order) encoding, and self-attention.
\cjx{
    The self-attention module is the core component, generating 
    \mtj{refined} attention feature \mtj{for its input feature} based on global context.
    First, self-attention takes the sum of input embedding and positional encoding as input, and
    computes three vectors for each word: \textit{query}, \textit{key} and \textit{value} through trained linear layers.
    Then, the attention weight between any two words can be obtained by matching (dot-producting) their {query} and {key} vectors.
    Finally, the attention feature is defined as the weighted sum of all {value} vectors with the attention weights. 
    Obviously, the output attention feature of each word is related to all input features, making it capable of learning the global context.
}
All operations of {Transformer} are parallelizable and order-independent.
In theory, it can replace the convolution operation in a convolutional neural network and has better versatility.
For more detailed introduction of self-attention, please refer to Section~\ref{sec:naivePCT}.


{
Inspired by the {Transformer}'s success in vision and NLP tasks, we propose a novel framework PCT for point cloud learning based on the principles of traditional Transformer.
\gmh{The key idea of PCT is using the inherent order invariance of Transformer to avoid the need to define the order of point cloud data and conduct feature learning through the attention mechanism.}
As shown in Figure~\ref{fig:attention}, the distribution of attention weights is highly related to part semantics, and it does not seriously attenuate with spatial distance.
}

{
Point clouds and natural language are rather different kinds of data, so our PCT framework must make several adjustments for this. These include:
\begin{itemize}
    \item
          \textbf{Coordinate-based input embedding module.}
          In Transformer, a positional encoding module is applied to represent the word order in nature language.
          This can distinguish the same word in different positions and reflect the positional relationships between words.
          However, point clouds do not have a fixed \gmh{order}.
          In our PCT framework, we merge the raw positional encoding and the input embedding into a coordinate-based input embedding module.
          It can generate distinguishable features, since each point has \textbf{unique} coordinates which represent its spatial position.
    \item
          \textbf{Optimized offset-attention module.}
          The offset-attention module approach we proposed is an effective upgrade over the original self-attention.
          It works by replacing the attention feature with the offset between the input of self-attention module and attention feature.
          This has two advantages.
          \cjx{Firstly, \mtj{the absolute coordinates of the same object can be completely different with rigid transformations.}
              Therefore, relative coordinates are generally more robust.
          }
          Secondly, the Laplacian matrix (the offset between degree matrix and adjacency matrix) has been proven to be very effective in graph convolution learning~\cite{gcn_2014}.
          From this perspective, we regard the point cloud as a graph with the ‘float’ adjacency matrix as the attention map.
          Also, the attention map in our work will be scaled with all the sum of each rows to 1.
          So the degree matrix can be understood as the identity matrix.
          Therefore, the offset-attention optimization process can be approximately understood as a Laplace process, which will be discuss detailed in Section~\ref{sec:offatt}.
          In addition, we have done sufficient comparative experiments, introduced in Section~\ref{sec:exp}, on offset-attention and self-attention to prove its effectiveness.
    \item
          \textbf{Neighbor embedding module.}
          Obviously, every word in a sentence contains basic semantic information.
          However, the \gmh{independent} input coordinates of the points are only weakly related to the semantic content.
          \gmh{Attention mechanism is effective in capturing global features, but it may ignore local geometric information which is also essential for point cloud learning.}
          To address this problem, we use a neighbor embedding strategy to improve upon point embedding.
          \gmh{It} also assists the attention module \gmh{by considering attention between local groups of points containing semantic information instead of individual points.}
\end{itemize}
}

\gmh{With the above adjustments, the PCT becomes more suitable for point cloud feature learning and achieves the state-of-the-art performance on shape classification, part segmentation and normal estimation tasks.
}

The main contributions of this paper are summarized as following:

\begin{enumerate}
    \item
          \gmh{We proposed a novel transformer based framework
              named PCT for point cloud learning, which is exactly suitable for unstructured, disordered point cloud data with irregular domain.}
    \item
          \gmh{We proposed offset-attention with implicit Laplace operator and normalization refinement which is inherently permutation-invariant and more suitable for point cloud learning compare to the original self-attention module in Transformer.}

    \item
          \gmh{Extensive experiments demonstrate that the PCT with explicit local context enhancement achieves state-of-the-art performance on shape classification, part segmentation and normal estimation tasks.}

\end{enumerate}

\section{Related Work}

\subsection{Transformer in NLP}
Bahdanau et al.~\cite{Bahdanau2015neural} proposed a neural machine translation method with an attention mechanism, in which attention weight is computed through the hidden state of an RNN.
Self-attention was proposed by Lin et al.~\cite{Lin2017selfatt} to visualize and interpret sentence embeddings.
Building on these, Vaswani et al.~\cite{Vaswani2017attention} proposed Transformer for machine translation; it is based solely on self-attention, without any recurrence or convolution operators.
Devlin et al.~\cite{Devlin2019bert} proposed bidirectional transformers~(BERT) approach, which is one of the most powerful models in the NLP field.
More lately, language learning networks such as XLNet~\cite{Yang2019xlnet}, Transformer-XL~\cite{Dai2019txl} and BioBERT~\cite{Lee2020biobert} have further extended the Transformer framework.

However, in natural language processing, the input is \gmh{in order}, and \gmh{word has} basic semantic, whereas point clouds are \gmh{unordered}, and individual points have no semantic meaning in general.

\subsection{Transformer for vision}
Many frameworks have introduced attention into vision tasks.
Wang et al.~\cite{Wang2017resatt} proposed a residual attention approach with stacked attention modules for image classification.
Hu et al.~\cite{Hu2018SENET} presented a novel spatial encoding unit, the SE block, whose idea was derived from the attention mechanism.
Zhang el al.~\cite{Zhang2019sagan} designed SAGAN, which uses self-attention for image generation.
There has also been an increasing trend to employ Transformer as a module to optimize neural networks.
Wu et al.~\cite{Wu2020visual} proposed visual transformers that apply Transformer to token-based images from feature maps for vision tasks.  Recently, Dosovitskiy~\cite{Dosovitskiy2020ViT}, proposed an image recognition network, ViT, based on patch encoding and Transformer, showing that with sufficient training data, Transformer provides better performance than a traditional convolutional neural network.
Carion et al.~\cite{Carion2020e2e} presented an end-to-end detection transformer that takes CNN features as input and generates bounding boxes with a Transformer encoder-decoder.

\gmh{Inspired by the local patch structures used in ViT and basic semantic information in language word, we present a neighbor embedding module that aggregates features from a point's local neighborhood, which can capture the local information and obtain semantic information.}

\subsection{Point-based deep learning}
PointNet~\cite{qi2016pointnet}  pioneered point cloud learning.
Subsequently, Qi et al. proposed PointNet++~\cite{qi2017pointnet++}, which uses query ball grouping and hierarchical PointNet to capture local structures.
Several subsequent \gmh{works} considered how to define convolution operations on point clouds.
One main approach is to convert a point cloud into a regular voxel array to allow convolution operations.
Tchapmi et al.~\cite{Tchapmi2017segcloud} proposed SEGCloud for pointwise segmentation.
It maps convolution features of 3D voxels to point clouds using trilinear interpolation and keeps global consistency through fully connected conditional random fields.
Atzmon et al~\cite{Atzmon2018point} present the PCNN framework with extension and restriction operators to map between point-based representation and voxel-based representation.
Volumetric convolution is performed on voxels for point feature extraction.
MCCNN by Hermosilla et al.~\cite{hermosilla2018monte} allows non-uniformly sampled point clouds; convolution is treated as a Monte Carlo integration problem.
Similarly, in PointConv proposed by Wu et al.~\cite{Wu2019pointconv},  3D convolution is performed through Monte Carlo estimation and importance sampling.

A different approach redefines convolution to operation on irregular point cloud data.
Li et al.~\cite{li2018pointcnn} introduce a point cloud convolution network, PointCNN, in which a $\chi$-transformation is trained to determine a 1D point order for convolution.
Tatarchenko et al.~\cite{tatarchenko2018tangent} proposed tangent convolution, which can learn surface geometric features from projected virtual tangent images.
SPG proposed by Landrieu et al.~\cite{landrieu2018large} divides the scanned scene into similar elements, and establishes a superpoint graph structure to learn contextual relationships between object parts.
Pan et al.~\cite{pan2018convolutional} use a parallel framework to extend  CNN from the conventional domain to a curved two-dimensional manifold.
However, it requires dense 3D gridded data as input so is unsuitable for 3D point clouds.
Wang et al.~\cite{wang2019dynamic} designed an EdgeConv operator for dynamic graphs, allowing point cloud learning by recovering local topology.

Various other methods also employ attention and Transformer.
Yan et al.~\cite{yan2020pointasnl} proposed PointASNL to deal with noise in point cloud processing, using a self-attention mechanism to update features for local groups of points.
Hertz et al.~\cite{Hertz2020pointgmm} proposed PointGMM for shape interpolation with both multi-layer perceptron (MLP)
splits and attentional splits.

Unlike the above methods, our PCT is based on Transformer rather than {using self-attention as an auxiliary module}. While a framework by Wang et al.~\cite{wang2019deep} uses Transformer to optimize point cloud registration, our PCT is a more general framework which can be used for various point cloud tasks.

\begin{figure*}[t]
    \centering
    \includegraphics[width=\textwidth]{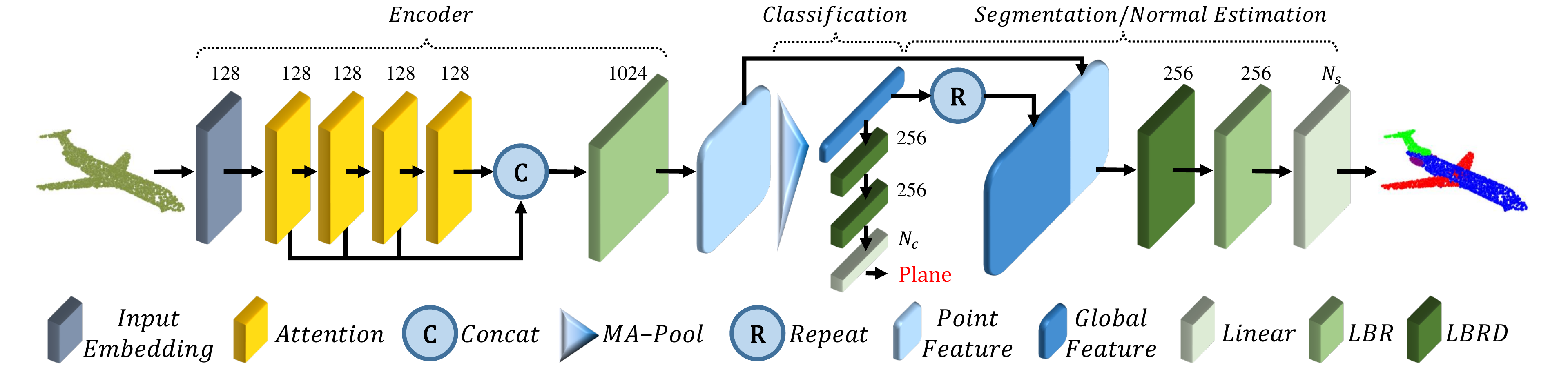}
    \caption{
    PCT architecture.
    {
    The encoder mainly comprises an \textit{Input Embedding} module and four stacked \textit{Attention} module.
    The decoder  mainly comprises multiple \textit{Linear} layers.
    Numbers above each module indicate its output channels.
    \textit{MA-Pool} concatenates \textit{Max-Pool} and \textit{Average-Pool}.
    \textit{LBR} combines \textit{Linear}, \textit{BatchNorm} and \textit{ReLU} layers.
    \textit{LBRD} means \textit{LBR} followed by a \textit{Dropout} layer.
    }
    }
    \label{fig:spct}
    \vspace{-1ex}
\end{figure*}

\section{Transformer for Point Cloud Representation}

{In this section, we first show how the point cloud representation learned by our PCT can be applied to various tasks of point cloud processing, \gmh{including} point cloud classification, part segmentation and normal estimation.
Thereafter, we detail the design of PCT.
We first introduce a na\"{i}ve version of PCT by directly applying the original Transformer~\cite{Vaswani2017attention} to point clouds.
We then explain full PCT with its special attention mechanism, and neighbor aggregation to provide enhanced local information.
}

\subsection{Point Cloud Processing with PCT}
\label{sec:tasks}

\textbf{Encoder.} The overall architecture of PCT is presented in Figure~\ref{fig:spct}.
PCT aims to transform (encode) the input points into a new higher dimensional feature space, which can characterize the semantic affinities between points as a basis for various point cloud processing tasks.
The encoder of PCT starts by embedding the input coordinates into a new feature space.
The embedded features are later fed into $4$ stacked attention module to learn a semantically rich and discriminative representation for each point, followed by a linear layer to generate the output feature.
Overall, the encoder of PCT shares almost the same philosophy of design as the original Transformer, except that the positional embedding is discarded, since the point’s coordinates already contains this information.
We refer the reader to~\cite{Vaswani2017attention} for details of the original NLP Transformer.

Formally, \mtj{given} 
\mtj{an input} point cloud $\mathcal{P}\in \mathbb{R}^{N\times d}$
with $N$ points each having $d$-dimensional feature description, \mtj{a $d_e$-dimensional embedded feature $\mathbf{F}_e \in \mathbf{R}^{N\times d_e}$ is first learned via the \emph{Input Embedding} module}.
%
The {point-wise} $d_o$-dimensional feature representation $\mathbf{F}_o\in \mathbb{R}^{N\times d_o}$ output by PCT is then formed by concatenating the attention output of each attention layer {through the feature dimension}, followed by a linear transformation:
\begin{align}
    \mathbf{F}_1   & = \mathrm{AT}^{1}(\mathbf{F}_e), \nonumber                                                    \\
    \mathbf{F}_{i} & = \mathrm{AT}^{i}(\mathbf{F}_{i-1}),\quad i=2,3,4, \nonumber                                  \\
    \mathbf{F}_o   & = \mathrm{concat}( \mathbf{F}_1, \mathbf{F}_2, \mathbf{F}_3, \mathbf{F}_4)\cdot \mathbf{W}_o,
\end{align}
where $\mathrm{AT}^i$ represents the $i$-th attention layer, each having the same output dimension as its input, and $\mathbf{W}_o$ is the weights of the linear layer.
\mtj{Various implementations of input embedding and attention will be explained later.}

%
{To extract} an effective
global feature vector {$\mathbf{F}_g$} representing the point cloud, we choose to concatenate the outputs from two pooling operators: a max-pooling (MP) and an average-pooling (AP) on the learned point-wise feature representation~\cite{wang2019dynamic}.

\textbf{Classification.} \mtj{The details of classification network using PCT is shown in Figure~\ref{fig:spct}.} To classify a point {cloud} $\mathcal{P}$ into $N_c$ object categories (e.g.\ desk, table, chair), we feed the global feature $\mathbf{F}_g$ to the classification decoder, 
\mtj{which} comprises two cascaded feed-forward neural networks {LBRs} (combining Linear, BatchNorm (BN) and ReLU layers) each with a dropout probability of $0.5$, finalized by a Linear layer to predict the final classification scores $\mathcal{C} \in \mathbb{R}^{N_c}$.
The class label of the point cloud is determined as the class with maximal score.

\textbf{Segmentation.} For the task of segmenting the point cloud into $N_s$ parts (e.g.\ table top, table legs; a part need not be contiguous),
we must predict a part label for each point,
we first concatenate the global feature $\mathbf{F}_g$ with {the point-wise features in} $\mathbf{F}_o$.
{To learn a common model for various kinds of objects, we also encode the one-hot object category vector as a $64$-dimensional feature and concatenate it with the global feature, following most other point cloud segmentation networks~\cite{qi2017pointnet++}.}
\mtj{As shown in Figure~\ref{fig:spct},}
the architecture of the segmentation network decoder is almost the same as that for the classification network, except that dropout is only performed on the first LBR.
We then predict the final {point-wise} segmentation scores $\mathcal{S} \in \mathbb{R}^{N\times N_s}$ for the input point cloud:
Finally, the part label of a point is also determined as the one with maximal score.

\textbf{Normal estimation.} {For the task of normal estimation,} we use the same architecture as in segmentation by setting $N_s = 3$, without the object category encoding, \mtj{and regard the output point-wise score as the predict normal}.

\begin{figure*}[t]
    \centering
    \includegraphics[width=\textwidth]{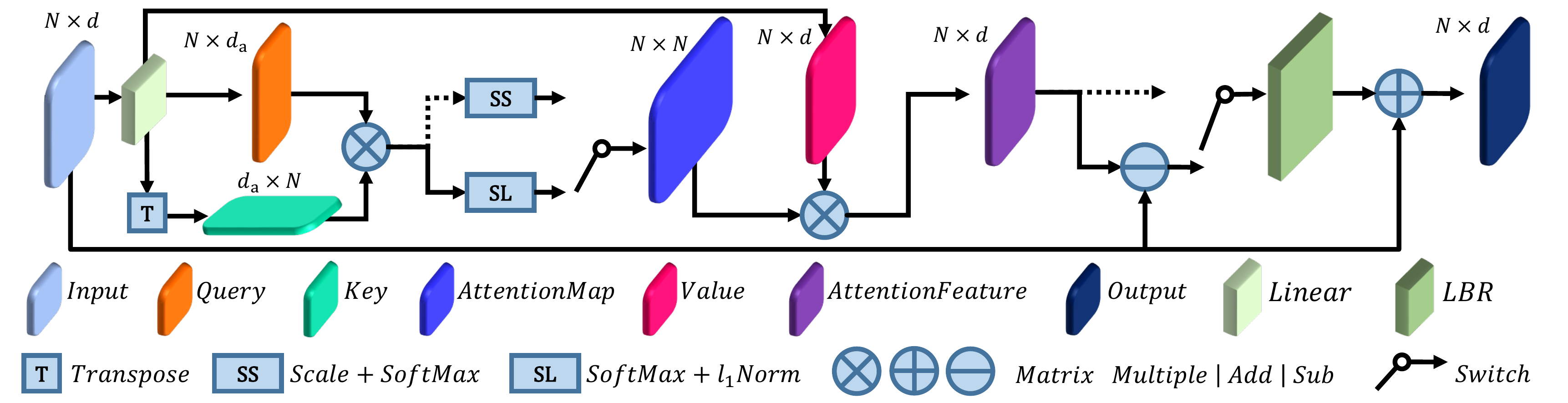}
    \caption{{
                Architecture of Offset-Attention.
                Numbers above tensors are numbers of dimensions $N$ and feature channels $D/D_a$, with switches showing alternatives of Self-Attention or Offset-Attention: dotted lines indicate Self-Attention branches.
            }
    }
    \label{fig:offset_att}
    \vspace{-2ex}
\end{figure*}

\subsection{Na\"{i}ve PCT}
\label{sec:naivePCT}

The simplest way to modify Transformer~\cite{Vaswani2017attention} for point cloud use is to treat the entire point cloud as a sentence and each point as a word, an approach we now explain.
\mtj{This na\"{i}ve PCT is achieved by implementing a coordinate-based point embedding  and instantiating the attention layer with the self-attention introduced in~\cite{Vaswani2017attention}.}

First, we consider a na\"{i}ve point embedding, {which ignores interactions between points.}
Like  word embedding in NLP, point embedding aims to place points closer in the embedding space if they are more semantically similar.
{Specifically,} we embed a point {cloud $\mathcal{P}$}
into a $d_e$-dimensional space $\mathbf{F}_e \in \mathbb{R}^{N\times d_e}$, {using} a shared \mtj{neural network comprising two cascaded LBRs, each with a $d_e$-dimensional output}. 
We {empirically} set $d_e=128$, a relatively small value, for  computational efficiency.
We simply use the point's 3D coordinates as its input feature description (i.e.\ $d_p=3$) (as doing so still outperforms other methods) but additional point-wise input information, such as point normals, could also be used.

{For the na\"{i}ve implementation of PCT, we adopt self-attention (SA) as introduced in the original Transformer~\cite{Vaswani2017attention}.}
Self-attention, also called intra-attention, is a mechanism that calculates {semantic affinities} between different items within a sequence of data.
{The architecture of the SA layer is depicted in Figure~\ref{fig:offset_att} by switching to the dotted data flows.}
{Following the terminology} in~\cite{Vaswani2017attention}, let
$\mathbf{Q,K,V}$ be the \emph{query, key} and \emph{value} matrices, respectively,
generated {by linear transformations of the input features $\mathbf{F}_{in} \in \mathbb{R}^{N\times d_e}$} 
as follows:
\begin{align}\label{eq:qkv}
    (\mathbf{Q,K,V})           & = \mathbf{F}_{in} \cdot (\mathbf{W}_q, \mathbf{W}_k, \mathbf{W}_v)\nonumber         \\
    \mathbf{Q,K}               & \in \mathbb{R}^{N\times d_a},\quad\mathbf{V} \in \mathbb{R}^{N\times d_e} \nonumber \\
    \mathbf{W}_q, \mathbf{W}_k & \in \mathbb{R}^{d_e \times d_a},\quad \mathbf{W}_v \in \mathbb{R}^{d_e \times d_e}
\end{align}
where $\mathbf{W}_q$, $\mathbf{W}_k$ and $\mathbf{W}_v$ are the shared learnable {linear transformation, and $d_a$ is the dimension of the query and key vectors.}
Note that $d_a$ may not be equal to $d_e$.
In this work, we set $d_a$ to be $d_e/4$ {for computational efficiency}.

First, we can use the query and key matrices to {calculate the attention weights} via the matrix dot-product:
\begin{equation}
    \Tilde{\mathbf{A}}=(\Tilde{\alpha})_{i,j} = \mathbf{Q} \cdot \mathbf{K}^{\mathrm{T}}.
\end{equation}
{These weights are then normalized {(denoted SS in Figure~\ref{fig:offset_att})} to give ${\mathbf{A}}=({\alpha})_{i,j}$:}
\begin{align}\label{eq:norm}
    \bar{\alpha}_{i,j} & =\frac{\Tilde{\alpha}_{i,j}}{\sqrt{d_a}}, \nonumber                                                                      \\
    {\alpha}_{i,j}     & = \mathrm{softmax}(\bar{\alpha}_{i,j}) = \frac{\exp{(\bar{\alpha}_{i,j})}}{\sum\limits_{k}{\exp{(\bar{\alpha}_{i,k}})}},
\end{align}
The self-attention output features $\mathbf{F}_{sa}$ are  the weighted sums of the value vector using the corresponding attention weights:
\begin{equation}\label{eq:sa}
    \mathbf{F}_{sa} = \mathbf{A}\cdot \mathbf{V}
\end{equation}

As the query, key and value matrices are determined by the shared corresponding {linear transformation} matrices and  the input feature $\mathbf{F}_{in}$, they are all order independent.
Moreover, $\mathrm{softmax}$ and weighted sum are both permutation-independent operators.
Therefore, {the whole self-attention process is}
permutation-invariant, {making it well-suited to the disordered, irregular domain presented by point clouds.}

    {Finally, the self-attention feature $\mathbf{F}_{sa}$ and the input feature $\mathbf{F}_{in}$, are further used to provide the output feature $\mathbf{F}_{out}$ for the whole SA layer through an LBR network:}
\begin{equation}\label{eq:oa}
    \mathbf{F}_{out}=\mathrm{SA}(\mathbf{F}_{in})=\mathrm{LBR}(\mathbf{F}_{sa})+\mathbf{F}_{in}.
\end{equation}

\subsection{Offset-Attention}
\label{sec:offatt}

{Graph convolution networks~\cite{gcn_2014} show the benefits of using a Laplacian matrix $\mathbf{L}=\mathbf{D}-\mathbf{E}$ to replace the adjacency matrix $\mathbf{E}$,
where $\mathbf{D}$ is the diagonal degree matrix.
{Similarly, we find that we can 
obtain better network performance if, when applying Transformer to point clouds,} we replace the original self-attention (SA) module with  an offset-attention (OA) module \mtj{to enhance our PCT}.}
As shown in Figure~\ref{fig:offset_att}, the offset-attention layer calculates the offset (difference) between the self-attention (SA) features
and the input features by element-wise subtraction.
{This offset feeds the LBR network in place of the SA feature used in the na\"{i}ve version.}
{Specifically, Equation~\ref{eq:sa} is modified to:}
\begin{align}\label{eq:oa}
    \mathbf{F}_{out} =\textrm{OA}(\mathbf{F}_{in})= & \textrm{LBR}(\mathbf{F}_{in} - \mathbf{F}_{sa}) + \mathbf{F}_{in}.
\end{align}

$\mathbf{F}_{in} - \mathbf{F}_{sa}$ is analogous to a discrete Laplacian operator, as we now show.
First, from Equations~\ref{eq:qkv} and~\ref{eq:sa}, the following holds:
\begin{align}
    \mathbf{F}_{in} - \mathbf{F}_{sa} & = \mathbf{F}_{in} - \mathbf{A} \mathbf{V} \nonumber                           \\
                                      & = \mathbf{F}_{in} - \mathbf{A} \mathbf{F}_{in} \mathbf{W}_v. \nonumber        \\
                                      & \approx \mathbf{F}_{in} - \mathbf{A} \mathbf{F}_{in} \nonumber                \\
                                      & = (\mathbf{I} - \mathbf{A})\mathbf{F}_{in} \approx \mathbf{L}\mathbf{F}_{in}.
\end{align}
Here, $\mathbf{W}_v$ is ignored since it is a weight matrix of the \textit{Linear layer}.
$\mathbf{I}$ is an identity matrix comparable to the diagonal degree matrix $D$ of the Laplacian matrix and $\mathbf{A}$ is the attention matrix comparable to the adjacency matrix $\mathbf{E}$.

\begin{figure*}[t!]
    \centering
    \includegraphics[width=\textwidth]{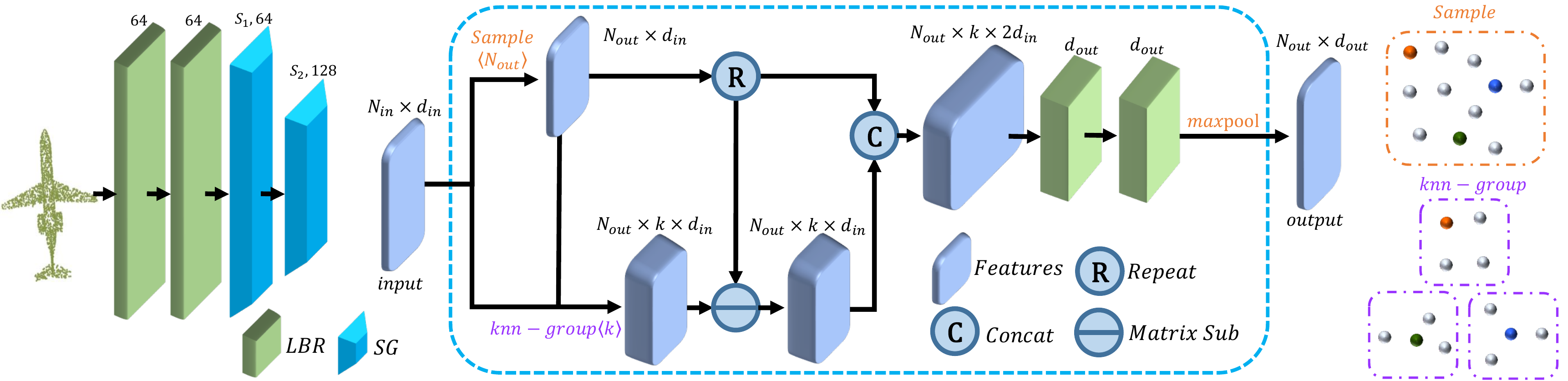}
    \caption{{
                Left: Neighbor Embedding architecture;
                Middle: SG Module with  $N_{in}$ input points,  $D_{in}$ input channels, $k$ neighbors, $N_{out}$ output sampled points and  $D_{out}$ output channels;
                Top-right: example of sampling (colored balls represent sampled points);
                Bottom-right: example of grouping with $k$-NN neighbors;
                Number above  LBR: number of output channels.
                Number above SG:  number of sampled points and its output channels.
            }
    }
    \label{fig:ne}
    \vspace{-2ex}
\end{figure*}

{In our enhanced version of PCT, we also refine the normalization  by modifying Equation~\ref{eq:norm} as follows:}
\begin{align}
    \bar{\alpha}_{i,j} & = \mathrm{softmax}(\Tilde{\alpha}_{i,j}) = \frac{\exp{(\Tilde{\alpha}_{i,j})}}{\sum\limits_{k}{\exp{(\Tilde{\alpha}_{k,j}})}},
    \nonumber                                                                                                                                           \\
    \alpha_{i,j}       & = \frac{{\bar{\alpha}_{i,j}}}{\sum\limits_{k}{{\bar{\alpha}_{i,k}}}}
\end{align}
Here, we use the $\mathrm{softmax}$ operator on the first dimension and an $l_1$-norm for the second dimension to normalize the attention map.
The traditional Transformer scales the first dimension by {$1/\sqrt{d_a}$} and uses $\mathrm{softmax}$ to normalize the second dimension.
However, our {offset-attention} sharpens the attention weights \gmh{and reduce the influence of noise}, which is beneficial for \gmh{downstream} tasks.
%
Figure~\ref{fig:attention} shows example offset attention maps.
It can be seen that the attention maps for different query points vary considerably, but are generally semantically meaningful.
\mtj{We refer to this refined PCT, i.e.\ with point embedding and OA layer, as simple PCT (SPCT) in the experiments.}

\subsection{Neighbor Embedding for Augmented Local Feature Representation}
\label{sec:ne}

PCT with point embedding is an effective {network} for extracting global features.
\gmh{However, it ignore the local neighborhood information which is also essential in point cloud learning.}
We draw upon the ideas of PointNet++~\cite{qi2017pointnet++} and DGCNN~\cite{wang2019dynamic} to design a local neighbor aggregation strategy, \emph{neighbor embedding}, to optimize the point embedding to augment PCT's ability of local feature extraction.
\cjx{
%
As shown in Figure~\ref{fig:ne}, neighbor embedding module comprises two {LBR} layers and two {SG (sampling and grouping)} layers.
The {LBR} layers act as the basis point embedding in Section~\ref{sec:naivePCT}.
{We use two cascaded SG layers to gradually enlarge the receptive field during feature aggregation, as is done in CNNs.}}
The {SG} layer {aggregates} features {from} the local neighbors {for each point} grouped by $k$-NN search  \gmh{using Euclidean distance}
during point cloud sampling.

{More specifically,} {assume that SG layer} takes a point cloud $\mathcal{P}$ with $N$ points and corresponding features $\mathbf{F}$ as input and outputs a sampled point cloud $\mathcal{P}_s$ with $N_s$ points and its corresponding aggregated features $\mathbf{F}_s$.
First, We adopt the farthest point sampling (FPS) algorithm~\cite{qi2017pointnet++} to {downsample} $\mathcal{P}$ to $\mathcal{P}_s$.
Then, for each sampled point $p \in \mathcal{P}_s$, let $\mathrm{knn}(p, \mathcal{P})$ be its $k$-nearest neighbors in $\mathcal{P}$.
We then compute the output feature $\mathbf{F}_s$ as follows:
\begin{align}
    \Delta\mathbf{F}(p)        & = \mathrm{concat}_{q\in \mathrm{knn}(p,\mathcal{P})} (\mathbf{F}(q) - \mathbf{F}(p)) \nonumber \\
    \widetilde{\mathbf{F}}_(p) & = \mathrm{concat}(\Delta\mathbf{F}(p), \mathop{\mathrm{RP}}(\mathbf{F}(p), k)) \nonumber       \\
    \mathbf{F}_s(p)            & = \mathrm{MP}(\mathrm{LBR}(\mathrm{LBR}(\widetilde{\mathbf{F}}(p))))
\end{align}
where $\mathbf{F}(p)$ is the input feature of point $p$, $\mathbf{F}_s(p)$ is the output feature of sampled point $p$, $\mathrm{MP}$ is the max-pooling operator,
and $\mathrm{RP}(\mathbf{x},k)$ is the operator for repeating a vector $\mathbf{x}$ $k$ times to form a matrix.
The idea of concatenating the feature among sampled point and its neighbors is drawn from \textit{EdgeConv}\cite{wang2019dynamic}.

We use different architectures
for the tasks of point cloud classification, segmentation {and normal estimation}.
For the {point cloud classification}, we only need to predict a global class for all points, so the sizes of the point cloud are decreased to 512 and 256 points within the two SG layer.

For point cloud segmentation {or normal estimation}, we need to determine point-wise part labels {or normal}, so the process above is only used for local feature extraction without reducing the point cloud size, which can {be achieved by setting} the output at each stage to still be of size $N$.

\section{Experiments}
\label{sec:exp}

We now evaluate the performance of {na\"{i}ve PCT (NPCT,  with point embedding and self-attention), {simple} PCT (SPCT, with point embedding and offset-attention) {and  full} PCT (with neighbor embedding and offset-attention)} {on two public datasets, ModelNet40~\cite{Wu2015modelnet} and ShapeNet~\cite{yi2016shapenet}}, giving a comprehensive comparison with other methods.
\gmh{The same soft cross-entropy loss function as} ~\cite{wang2019dynamic} and the stochastic gradient descent~(SGD) optimizer
with momentum 0.9 were adopted for {training} in each case.
Other training parameters, including the learning rate, batch size and input format, were particular to each specific dataset and are given later.

\begin{table}[t!]
    \centering
    \caption{Comparison with state-of-the-art methods on the ModelNet40 classification dataset. Accuracy means overall accuracy.
        All results quoted are taken from the cited papers.
        P = points, N = normals.}
    \begin{tabular}{l|lrr}
        \hline
        \textbf{Method}                         & \textbf{input} & \textbf{\#points} & \textbf{Accuracy} \\
        \hline
        PointNet~\cite{qi2016pointnet}          & P              & 1k                & 89.2\%            \\
        
        A-SCN~\cite{Xie_2018_CVPR} & P & 1k & 89.8  \%            \\

        SO-Net~\cite{li2018sonet}               & P, N           & 2k                & 90.9\%            \\
        Kd-Net~\cite{Klokov2017kdnet}           & P              & 32k               & 91.8\%            \\
        PointNet++~\cite{qi2017pointnet++}      & P              & 1k                & 90.7\%            \\
        PointNet++~\cite{qi2017pointnet++}      & P, N           & 5k                & 91.9\%            \\
        PointGrid~\cite{Le2018Pointgrid}        & P              & 1k                & 92.0\%            \\
        PCNN~\cite{Atzmon2018point}             & P              & 1k                & 92.3\%            \\
        PointWeb~\cite{Zhao2019pointweb}        & P              & 1k                & 92.3\%            \\
        PointCNN~\cite{li2018pointcnn}          & P              & 1k                & 92.5\%            \\
        PointConv~\cite{Wu2019pointconv}        & P, N           & 1k                & 92.5\%            \\
        A-CNN~\cite{Komarichev2019acnn}         & P, N           & 1k                & 92.6\%            \\
        P2Sequence~\cite{liu2019Point2Sequence} & P              & 1k                & 92.6\%            \\
        KPConv~\cite{Thomas2019kpconv}          & P              & 7k                & 92.9\%            \\
        DGCNN~\cite{wang2019dynamic}            & P              & 1k                & 92.9\%            \\
        RS-CNN~\cite{Liu2019rscnn}              & P              & 1k                & 92.9\%            \\
        PointASNL~\cite{yan2020pointasnl}       & P              & 1k                & 92.9\%            \\
        \hline

        NPCT                                    & P              & 1k                & 91.0\%            \\
        SPCT                                    & P              & 1k                & 92.0\%            \\
        PCT                                     & P              & 1k                & \textbf{93.2\%}   \\

        \hline
    \end{tabular}
    \label{Tab.ModelNet40.classification}
    \vspace{-2ex}
\end{table}

\subsection{Classification on ModelNet40 dataset}
\label{sec:exp:cls}

{ModelNet40\cite{Wu2015modelnet} contains 12,311 CAD models in 40 object categories; it is widely used in point cloud shape classification and surface normal estimation benchmarking. For a fair comparison, we used the official split with 9,843 objects for training and 2,468 for evaluation. The same sampling strategy as used in PointNet~\cite{qi2016pointnet} was adopted to uniformly sample each object to 1,024 points.}
During training, a random translation in $[-0.2,0.2]$, a random anisotropic scaling in $[0.67, 1.5]$
and a random input dropout
were applied to augment the input data.
During testing, no data augmentation or voting methods were used.
\mtj{For all the three models,}
the mini-batch sizes were {32},
250 training epochs were used and the initial learning rates were {0.01}, with a cosine annealing schedule to adjust the learning rate at every epoch.

{Experimental results are shown in Table~\ref{Tab.ModelNet40.classification}}.
Compared to PointNet and NPCT, SPCT makes a 2.8\% and 1.0\% improvement respectively.
PCT achieves the best result of $93.2\%$  {overall} accuracy.
Note that our network currently does not consider normals as inputs which could in principle further {improve} network performance.

\begin{table}[ht]
    \centering
    \caption{Normal estimation average cosine-distance error on ModelNet40 dataset.}
    \begin{tabular}{l|rr}
        \hline
        \textbf{Method}                   & \textbf{\#Points} & \textbf{Error} \\
        \hline
        PointNet\cite{qi2016pointnet}     & 1k                & 0.47           \\
        PointNet++\cite{qi2017pointnet++} & 1k                & 0.29           \\
        PCNN \cite{Atzmon2018point}       & 1k                & 0.19           \\
        RS-CNN \cite{Liu2019rscnn}        & 1k                & 0.15           \\
        \hline
        NPCT                              & 1k                & 0.24           \\
        SPCT                              & 1k                & 0.23           \\
        PCT                               & 1k                & \textbf{0.13}  \\
        \hline
    \end{tabular}
    \label{Tab.NormalEstimation}
    \vspace{-2ex}
\end{table}

\subsection{Normal estimation on ModelNet40 dataset}
The surface normal estimation is to determine the
normal direction at each point. Estimating surface normal has wide applications in e.g.\ rendering. The task is challenging because it requires the approach to understand the shapes completely
for dense regression.
We again used ModelNet40 as a benchmark, and used \gmh{average} cosine distance to measure the difference between ground truth and predicted normals.
\mtj{For all the three models, a batch size of {32},
    200 training epochs were used.
    The initial learning rates were also set as {0.01}, with a cosine annealing schedule used to adjust learning rate every epoch.}
As indicated in Table~\ref{Tab.NormalEstimation},
both our NPCT and SPCT make a significant improvement compared with PointNet
and PCT achieves the lowest average cosine distance.


\subsection{Segmentation task on ShapeNet dataset}
Point cloud segmentation is a challenging task which aims to divide a 3D model into {multiple meaningful parts}.
We performed an experimental evaluation on the ShapeNet Parts dataset~\cite{yi2016shapenet}, which contains 16,880 3D models with a training to testing split of 14,006 to 2,874.
It has 16 object categories and 50 part labels; each instance contains no fewer than two parts.
{Following PointNet~\cite{qi2016pointnet}}, all models were downsampled to {2,048} points, retaining point-wise part annotation.
During  training, random translation in $[-0.2,0.2]$, and random anisotropic scaling in $[0.67, 1.5]$
were applied to augment the input data.
During testing, we used a multi-scale testing strategy,
where the scales are set in {$[0.7, 1.4]$ with a step of $0.1$.}
\mtj{For all the three models, the batch size, training epochs and the learning rates were set the same as the training of normal estimation task.}

\begin{figure*}[!t]
    \centering
    \includegraphics[width=\textwidth]{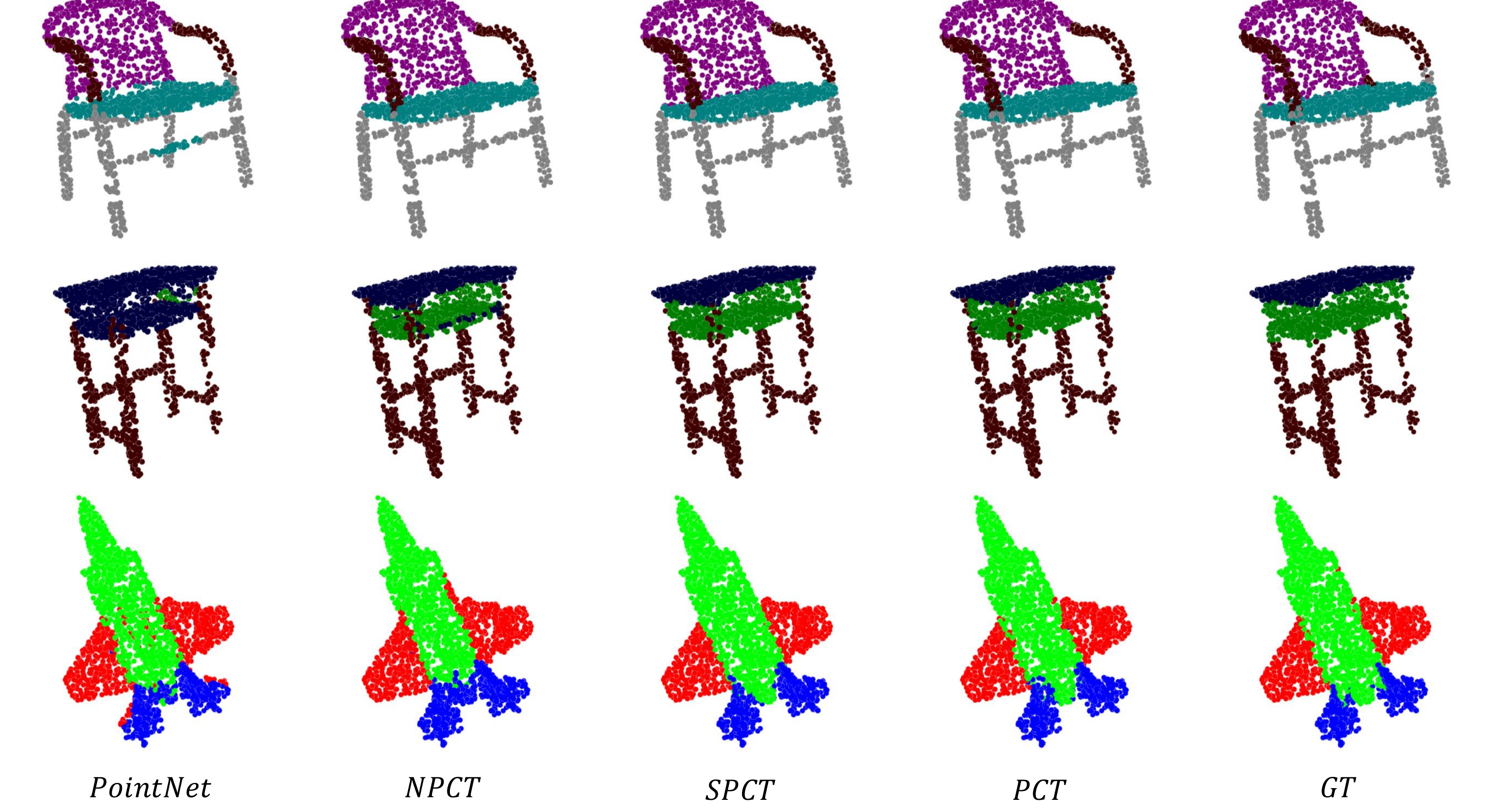}
    \caption{
        Segmentations from PointNet, NPCT, SPCT, PCT dnd Ground Truth(GT).
    }
    \label{fig.shapenet_example}
    \vspace{-2ex}
\end{figure*}

Table~\ref{Tab.ShapeNet} shows the class-wise segmentation results.
The evaluation metric used is  part-average Intersection-over-Union, and is given both overall and for each object category.
{The results show that our SPCT makes an improvement of $2.1\%$ and $0.6\%$ over PointNet and NPCT respectively.
PCT achieves the best results with $86.4\%$ part-average Intersection-over-Union.}
Figure~\ref{fig.shapenet_example} shows further segmentation examples provided by PointNet, NPCT, SPCT and PCT.

\begin{table*}[t!]
    \centering
    \caption{Comparison on the ShaperNet part segmentation dataset. pIoU means part-average Intersection-over-Union. All results quoted are taken from the cited papers.}
    \setlength{\tabcolsep}{1.9pt}
    \begin{tabular}{l|c|cccccccccccccccc}
        \hline
        \textbf{Method}                         & {pIoU}        & \tabincell{c}{{air}-\\{plane}} & {bag}         & {cap}         & {car}         & {chair}       & \tabincell{c}{{ear}-\\{phone}} & {guitar}      & {knife}       & {lamp}        & {laptop}      & \tabincell{c}{motor-                                                                                 \\bike} & {mug} & {pistol} & {rocket} & \tabincell{c}{{skate}-\\{board}}& {table} \\
        \hline
        PointNet~\cite{qi2016pointnet}          & 83.7          & 83.4                                                 & 78.7          & 82.5          & 74.9          & 89.6          & 73.0                                                 & 91.5          & 85.9          & 80.8          & 95.3          & 65.2                 & 93.0          & 81.2          & 57.9          & 72.8          & 80.6          \\
        Kd-Net~\cite{Klokov2017kdnet}           & 82.3          & 80.1                                                 & 74.6          & 74.3          & 70.3          & 88.6          & 73.5                                                 & 90.2          & 87.2          & 81.0          & 94.9          & 57.4                 & 86.7          & 78.1          & 51.8          & 69.9          & 80.3          \\

        SO-Net~\cite{li2018sonet}               & 84.9          & 82.8                                                 & 77.8          & 88.0          & 77.3          & 90.6          & 73.5                                                 & 90.7          & 83.9          & 82.8          & 94.8          & 69.1                 & 94.2          & 80.9          & 53.1          & 72.9          & 83.0          \\
        PointNet++~\cite{qi2017pointnet++}      & 85.1          & 82.4                                                 & 79.0          & 87.7          & 77.3          & 90.8          & 71.8                                                 & 91.0          & 85.9          & 83.7          & 95.3          & 71.6                 & 94.1          & 81.3          & 58.7          & 76.4          & 82.6          \\
        PCNN~\cite{Atzmon2018point}             & 85.1          & 82.4                                                 & 80.1          & 85.5          & 79.5          & 90.8          & 73.2                                                 & 91.3          & 86.0          & 85.0          & 95.7          & 73.2                 & 94.8          & 83.3          & 51.0          & 75.0          & 81.8          \\

        DGCNN~\cite{wang2019dynamic}            & 85.2          & 84.0                                                 & 83.4          & 86.7          & 77.8          & 90.6          & 74.7                                                 & 91.2          & 87.5          & 82.8          & 95.7          & 66.3                 & 94.9          & 81.1          & 63.5          & 74.5          & 82.6          \\
        P2Sequence~\cite{liu2019Point2Sequence} & 85.2          & 82.6                                                 & 81.8          & 87.5          & 77.3          & 90.8          & 77.1                                                 & 91.1          & 86.9          & 83.9          & 95.7          & 70.8                 & 94.6          & 79.3          & 58.1          & 75.2          & 82.8          \\
        PointConv~\cite{Wu2019pointconv}        & 85.7          & -                                                    & -             & -             & -             & -             & -                                                    & -             & -             & -             & -             & -                    & -             & -             & -             & -             & -             \\
        PointCNN~\cite{li2018pointcnn}          & 86.1          & 84.1                                                 & \textbf{86.5} & 86.0          & 80.8          & 90.6          & 79.7                                                 & \textbf{92.3} & \textbf{88.4} & 85.3          & \textbf{96.1} & \textbf{77.2}        & 95.2          & \textbf{84.2} & \textbf{64.2} & \textbf{80.0} & 83.0          \\
        PointASNL~\cite{yan2020pointasnl}       & 86.1          & 84.1                                                 & 84.7          & 87.9          & 79.7          & \textbf{92.2} & 73.7                                                 & 91.0          & 87.2          & 84.2          & 95.8          & 74.4                 & 95.2          & 81.0          & 63.0          & 76.3          & 83.2          \\
        RS-CNN~\cite{Liu2019rscnn}              & 86.2          & 83.5                                                 & 84.8          & 88.8          & 79.6          & 91.2          & \textbf{81.1}                                        & 91.6          & \textbf{88.4} & 86.0          & 96.0          & 73.7                 & 94.1          & 83.4          & 60.5          & 77.7          & 83.6          \\
        \hline
        \textbf{NPCT}                           & 85.2          & 83.2                                                 & 74.5          & 86.7          & 76.8          & 90.7          & 75.4                                                 & 91.1          & 87.3          & 84.5          & 95.7          & 65.2                 & 93.7          & 82.7          & 56.9          & 73.8          & 83.0          \\
        \textbf{SPCT}                           & 85.8          & 84.5                                                 & 83.5          & 85.9          & 78.7          & 90.9          & 75.1                                                 & 92.1          & 87.0          & 85.0          & 95.9          & 69.6                 & 94.5          & 82.2          & 61.4          & 76.0          & 83.0          \\
        \textbf{PCT}                            & \textbf{86.4} & \textbf{85.0}                                        & 82.4          & \textbf{89.0} & \textbf{81.2} & 91.9          & 71.5                                                 & 91.3          & 88.1          & \textbf{86.3} & 95.8          & 64.6                 & \textbf{95.8} & 83.6          & 62.2          & 77.6          & \textbf{83.7} \\

        \hline
    \end{tabular}
    \label{Tab.ShapeNet}
    \vspace{-1ex}
\end{table*}

\subsection{Semantic segmentation task on S3DIS dataset}

The S3DIS is a indoor scene dataset for point cloud semantic segmentation. It contains 6 areas and 271 rooms. Each point in the dataset is divided into 13 categories. For fair comparison, we use the same data processing method as~\cite{qi2016pointnet}. Table~\ref{Tab.S3DIS} shows that our PCT achieves superior performance compared to the previous methods.

\begin{table*}[t!]
    \centering
    \caption{Comparison on the S3DIS semantic segmentation dataset tested on Area5.}
    \setlength{\tabcolsep}{1.9pt}
    \begin{tabular}{l|cc|ccccccccccccc}
        \hline
        \textbf{Method}        & {mAcc}                 & {mIoU}        & \tabincell{c} {{ceil}-\\{ing}} & {floor}         & {wall}         & {beam}         & {column}       & {window} & {door}      & {chair}       & {table}        & \tabincell{c}{{book}-\\{case}} & {sofa} & {board} & {clutter}  \\
        \hline
        PointNet~\cite{qi2016pointnet}          &48.98 & 41.09 &88.80 &97.33 &69.80 &0.05 & 3.92 &46.26 &10.76 &58.93 &52.61 &5.85 & 40.28 & 26.38 & 33.22          \\

        SEGCloud~\cite{Tchapmi2017segcloud} &
         57.35 & 48.92 & 90.06 & 96.05 & 69.86 & 0.00 & 18.37 & 38.35 & 23.12 & 70.40 & 75.89 & 40.88 & 58.42 & 12.96 & 41.60
        \\

        DGCNN~\cite{wang2019dynamic}            & 84.10          & 56.10       & -   & -   & -   & -   & -   & -   & -   & -   & -   & -   & -   & -   & -                            \\

        PointCNN~\cite{li2018pointcnn}          & 63.86 & 57.26 & 92.31 & 98.24 & 79.41 & 0.00 & 17.60 & 22.77 & 62.09 & 74.39 & 80.59 & 31.67 & 66.67 & 62.05 & 56.74       \\

        SPG ~\cite{landrieu2018large}          &  66.50 & 58.04 & 89.35 & 96.87 & 78.12 & 0.00 & 42.81 & 48.93 & 61.58 & 84.66 & 75.41 & 69.84 & 52.60 & 2.10 & 52.22
        \\

        PCNN~\cite{Atzmon2018point}             & 67.01 & 58.27 & 92.26 & 96.20 & 75.89 & 0.27 & 5.98 & 69.49 & 63.45 & 66.87 & 65.63 & 47.28 & 68.91 & 59.10 & 46.22
        \\

        PointWeb~\cite{Zhao2019pointweb}          &  66.64 & 60.28 & 91.95 & 98.48 & 79.39 & 0.00 & 21.11 & 59.72 & 34.81 & 76.33 & 88.27 & 46.89 & 69.30 & 64.91 & 52.46      \\

        \hline
        \textbf{PCT}                            & \textbf{67.65} & \textbf{61.33}                                        & 92.54         & 98.42 & 80.62 & 0.00         & 19.37                 & 61.64          & 48.00         & 76.58 & 85.20        & 46.22                 & 67.71 & 67.93         & 52.29      \\

        \hline
    \end{tabular}
    \label{Tab.S3DIS}
    \vspace{-1ex}
\end{table*}

\begin{table}[tbp]
    \centering
    \caption{Computational resource requirements.}
    \begin{tabular}{l|rrr}
        \hline
        \textbf{Method}                          & {\#Params}     & \tabincell{c}{\#FLOPs} & {Accuracy}      \\
        \hline
        PointNet \cite{qi2016pointnet}           & 3.47M          & \textbf{0.45G}         & 89.2\%          \\
        PointNet++(SSG)  \cite{qi2017pointnet++} & 1.48M          & 1.68G                  & 90.7\%          \\
        PointNet++(MSG) \cite{qi2017pointnet++}  & 1.74M          & 4.09G                  & 91.9\%          \\
        DGCNN   \cite{wang2019dynamic}           & 1.81M          & 2.43G                  & 92.9\%          \\
        \hline
        NPCT                                     & \textbf{1.36M} & 1.80G                   & 91.0\%          \\
        SPCT                                     & \textbf{1.36M} & 1.82G                   & 92.0\%          \\
        PCT                                      & 2.88M          & 2.32G                  & \textbf{93.2\%} \\
        \hline
    \end{tabular}
    \label{Tab.params}
    \vspace{-2ex}
\end{table}

\begin{table}[t!]
    \centering
    \caption{Comparison on the ModelNet40 classification dataset. PCT-2L means PCT with 2 layer neighbor embedding and PCT-3L means PCT with 3 layer neighbor embedding.  Accuracy means overall accuracy.
        P = points.}
    \begin{tabular}{l|lrr}
        \hline
        \textbf{Method} & \textbf{input} & \textbf{\#points} & \textbf{Accuracy} \\
        \hline
        PCT-2L          & P              & 1k                & 93.2\%            \\
        \hline
        PCT-3L          & P              & 1k                & \textbf{93.4\%}   \\

        \hline
    \end{tabular}
    \label{Tab.3_layer_cls}
    \vspace{-2ex}
\end{table}

\begin{table*}[t!]
    \centering
    \caption{Comparison on the ShaperNet part segmentation dataset. pIoU means part-average Intersection-over-Union.PCT-2L means PCT with 2 layer neighbor embedding and PCT-3L means PCT with 3 layer neighbor embedding.}
    \setlength{\tabcolsep}{1.9pt}
    \begin{tabular}{l|c|cccccccccccccccc}
        \hline
        \textbf{Method} & {pIoU}        & \tabincell{c}{{air}-\\{plane}} & {bag}         & {cap}         & {car}         & {chair}       & \tabincell{c}{{ear}-\\{phone}} & {guitar}      & {knife}       & {lamp}        & {laptop}      & \tabincell{c}{motor-                                                                                 \\bike} & {mug} & {pistol} & {rocket} & \tabincell{c}{{skate}-\\{board}}& {table} \\
        \hline
        \textbf{PCT-2L} & 86.4          & 85.0                                                 & 82.4          & 89.0          & \textbf{81.2} & \textbf{91.9} & 71.5                                                 & 91.3          & \textbf{88.1} & \textbf{86.3} & 95.8          & 64.6                 & \textbf{95.8} & \textbf{83.6} & \textbf{62.2} & \textbf{77.6} & \textbf{83.7} \\
        \textbf{PCT-3L} & \textbf{86.6} & \textbf{85.3}                                        & \textbf{84.5} & \textbf{89.4} & 81.0          & 91.7          & \textbf{78.6}                                        & \textbf{91.5} & 87.5          & 85.8          & \textbf{96.0} & \textbf{70.6}        & 95.6          & 82.8          & 60.9          & 76.6          & \textbf{83.7} \\

        \hline

        \hline
    \end{tabular}
    \label{Tab.3_layer_seg}
    \vspace{-1ex}
\end{table*}

\subsection{Computational requirements analysis}
We now consider the computational requirements of NPCT, SPCT, PCT and several other methods {by comparing the floating point operations required (FLOPs) and number of parameters (Params)  in Table~\ref{Tab.params}.
        SPCT has the lowest memory requirements with only 1.36M parameters
        and also puts a low load on the processor of only 1.82 GFLOPs, yet delivers highly accurate results.
        These characteristics make it suitable for deployment on a mobile device.
        PCT has best performance, yet modest computational and memory requirements.} If we pursue higher performance and ignore the amount of calculation and parameters, we can add a neighbor embedding layer in the input embedding module. The results of 3-Layer embedding PCT are shown in Table~\ref{Tab.3_layer_cls} and \ref{Tab.3_layer_seg}.


\section{Conclusion}
In this paper, we propose a permutation-invariant point cloud transformer, which is suitable for learning on unstructured point clouds with irregular domain.
The proposed offset-attention and normalization mechanisms help to make our PCT effective.
Experiments show that PCT has good semantic feature learning capability, and achieves state-of-the-art performance on several tasks, particularly
shape classification, part segmentation and normal estimation.

Transformer has already revealed powerful capabilities given large amounts of training data.
\cjx{At present, the available point cloud datasets are very limited compared to image.
    In  future, we will train it
    on larger datasets and study its advantages and disadvantages with respect to other popular frameworks.
    Besides, the encoder-decoder structure of Transformer support more complex tasks, such as point cloud generation and completion.
    We will extend the PCT to further applications.
}

{\small
    \bibliographystyle{ieee_fullname}
    \bibliography{ref}

\begin{thebibliography}{10}\itemsep=-1pt

\bibitem{Atzmon2018point}
Matan Atzmon, Haggai Maron, and Yaron Lipman.
\newblock Point convolutional neural networks by extension operators.
\newblock {\em {ACM} Transactions on Graphics}, 37(4):71:1--71:12, 2018.

\bibitem{Bahdanau2015neural}
Dzmitry Bahdanau, Kyunghyun Cho, and Yoshua Bengio.
\newblock Neural machine translation by jointly learning to align and
  translate.
\newblock In {\em International Conference on Learning Representations}, 2015.

\bibitem{gcn_2014}
Joan Bruna, Wojciech Zaremba, Arthur Szlam, and Yann LeCun.
\newblock Spectral networks and locally connected networks on graphs.
\newblock In Yoshua Bengio and Yann LeCun, editors, {\em International
  Conference on Learning Representations}, 2014.

\bibitem{Carion2020e2e}
Nicolas Carion, Francisco Massa, Gabriel Synnaeve, Nicolas Usunier, Alexander
  Kirillov, and Sergey Zagoruyko.
\newblock {End-to-End} object detection with transformers.
\newblock {\em CoRR}, abs/2005.12872, 2020.

\bibitem{Dai2019txl}
Zihang Dai, Zhilin Yang, Yiming Yang, Jaime~G. Carbonell, Quoc~Viet Le, and
  Ruslan Salakhutdinov.
\newblock Transformer-xl: Attentive language models beyond a fixed-length
  context.
\newblock In Anna Korhonen, David~R. Traum, and Llu{\'{\i}}s M{\`{a}}rquez,
  editors, {\em Association for Computational Linguistics}, pages 2978--2988.
  Association for Computational Linguistics, 2019.

\bibitem{Devlin2019bert}
Jacob Devlin, Ming{-}Wei Chang, Kenton Lee, and Kristina Toutanova.
\newblock {BERT:} pre-training of deep bidirectional transformers for language
  understanding.
\newblock In Jill Burstein, Christy Doran, and Thamar Solorio, editors, {\em
  North American Chapter of the Association for Computational Linguistics:
  Human Language Technologies}, pages 4171--4186. Association for Computational
  Linguistics, 2019.

\bibitem{Dosovitskiy2020ViT}
Alexey Dosovitskiy, Lucas Beyer, Alexander Kolesnikov, Dirk Weissenborn,
  Xiaohua Zhai, Thomas Unterthiner, Mostafa Dehghani, Matthias Minderer, Georg
  Heigold, Sylvain Gelly, Jakob Uszkoreit, and Neil Houlsby.
\newblock An image is worth 16x16 words: Transformers for image recognition at
  scale.
\newblock {\em CoRR}, abs/2010.11929, 2020.

\bibitem{hermosilla2018monte}
Pedro Hermosilla, Tobias Ritschel, Pere{-}Pau V{\'{a}}zquez, {\`{A}}lvar
  Vinacua, and Timo Ropinski.
\newblock Monte carlo convolution for learning on non-uniformly sampled point
  clouds.
\newblock {\em {ACM} Transactions on Graphics}, 37(6):235:1--235:12, 2018.

\bibitem{Hertz2020pointgmm}
Amir Hertz, Rana Hanocka, Raja Giryes, and Daniel Cohen{-}Or.
\newblock {PointGMM}: {A} neural {GMM} network for point clouds.
\newblock In {\em {IEEE/CVF} Conference on Computer Vision and Pattern
  Recognition}, pages 12051--12060. {IEEE}, 2020.

\bibitem{Hu2018SENET}
Jie Hu, Li Shen, and Gang Sun.
\newblock Squeeze-and-excitation networks.
\newblock In {\em {IEEE} Conference on Computer Vision and Pattern
  Recognition}, pages 7132--7141. {IEEE} Computer Society, 2018.

\bibitem{Klokov2017kdnet}
Roman Klokov and Victor~S. Lempitsky.
\newblock Escape from cells: Deep kd-networks for the recognition of 3d point
  cloud models.
\newblock In {\em {IEEE} International Conference on Computer Vision}, pages
  863--872. {IEEE} Computer Society, 2017.

\bibitem{Komarichev2019acnn}
Artem Komarichev, Zichun Zhong, and Jing Hua.
\newblock {A-CNN:} annularly convolutional neural networks on point clouds.
\newblock In {\em {IEEE} Conference on Computer Vision and Pattern
  Recognition}, pages 7421--7430. Computer Vision Foundation / {IEEE}, 2019.

\bibitem{landrieu2018large}
Lo{\"{\i}}c Landrieu and Martin Simonovsky.
\newblock Large-scale point cloud semantic segmentation with superpoint graphs.
\newblock In {\em {IEEE} Conference on Computer Vision and Pattern
  Recognition}, pages 4558--4567. {IEEE} Computer Society, 2018.

\bibitem{Le2018Pointgrid}
Truc Le and Ye Duan.
\newblock Pointgrid: {A} deep network for 3d shape understanding.
\newblock In {\em {IEEE} Conference on Computer Vision and Pattern
  Recognition}, pages 9204--9214. {IEEE} Computer Society, 2018.

\bibitem{Lee2020biobert}
Jinhyuk Lee, Wonjin Yoon, Sungdong Kim, Donghyeon Kim, Sunkyu Kim, Chan~Ho So,
  and Jaewoo Kang.
\newblock Biobert: a pre-trained biomedical language representation model for
  biomedical text mining.
\newblock {\em Bioinformatics}, 36(4):1234--1240, 2020.

\bibitem{li2018sonet}
Jiaxin Li, Ben~M. Chen, and Gim~Hee Lee.
\newblock So-net: Self-organizing network for point cloud analysis.
\newblock In {\em {IEEE} Conference on Computer Vision and Pattern
  Recognition}, pages 9397--9406. {IEEE} Computer Society, 2018.

\bibitem{li2018pointcnn}
Yangyan Li, Rui Bu, Mingchao Sun, Wei Wu, Xinhan Di, and Baoquan Chen.
\newblock {PointCNN}: Convolution on x-transformed points.
\newblock In {\em Advances in Neural Information Processing Systems}, pages
  828--838, 2018.

\bibitem{Lin2017selfatt}
Zhouhan Lin, Minwei Feng, C{\'{\i}}cero~Nogueira dos Santos, Mo Yu, Bing Xiang,
  Bowen Zhou, and Yoshua Bengio.
\newblock A structured self-attentive sentence embedding.
\newblock In {\em International Conference on Learning Representations}.
  OpenReview.net, 2017.

\bibitem{liu2019Point2Sequence}
Xinhai Liu, Zhizhong Han, Yu{-}Shen Liu, and Matthias Zwicker.
\newblock Point2sequence: Learning the shape representation of 3d point clouds
  with an attention-based sequence to sequence network.
\newblock In {\em {AAAI} Conference on Artificial Intelligence}, pages
  8778--8785. {AAAI} Press, 2019.

\bibitem{Liu2019rscnn}
Yongcheng Liu, Bin Fan, Shiming Xiang, and Chunhong Pan.
\newblock Relation-shape convolutional neural network for point cloud analysis.
\newblock In {\em {IEEE} Conference on Computer Vision and Pattern
  Recognition}, pages 8895--8904. Computer Vision Foundation / {IEEE}, 2019.

\bibitem{qi2016pointnet}
Charles~Ruizhongtai Qi, Hao Su, Kaichun Mo, and Leonidas~J. Guibas.
\newblock Pointnet: Deep learning on point sets for 3d classification and
  segmentation.
\newblock In {\em {IEEE} Conference on Computer Vision and Pattern
  Recognition}, pages 77--85. {IEEE} Computer Society, 2017.

\bibitem{qi2017pointnet++}
Charles~Ruizhongtai Qi, Li Yi, Hao Su, and Leonidas~J. Guibas.
\newblock Pointnet++: Deep hierarchical feature learning on point sets in a
  metric space.
\newblock In {\em Advances in Neural Information Processing Systems}, pages
  5099--5108, 2017.

\bibitem{tatarchenko2018tangent}
Maxim Tatarchenko, Jaesik Park, Vladlen Koltun, and Qian{-}Yi Zhou.
\newblock Tangent convolutions for dense prediction in 3d.
\newblock In {\em {IEEE} Conference on Computer Vision and Pattern
  Recognition}, pages 3887--3896. {IEEE} Computer Society, 2018.

\bibitem{Tchapmi2017segcloud}
Lyne~P. Tchapmi, Christopher~B. Choy, Iro Armeni, JunYoung Gwak, and Silvio
  Savarese.
\newblock {SEGCloud}: Semantic segmentation of 3d point clouds.
\newblock In {\em International Conference on 3D Vision}, pages 537--547.
  {IEEE} Computer Society, 2017.

\bibitem{Thomas2019kpconv}
Hugues Thomas, Charles~R. Qi, Jean{-}Emmanuel Deschaud, Beatriz Marcotegui,
  Fran{\c{c}}ois Goulette, and Leonidas~J. Guibas.
\newblock Kpconv: Flexible and deformable convolution for point clouds.
\newblock In {\em {IEEE/CVF} International Conference on Computer Vision},
  pages 6410--6419. {IEEE}, 2019.

\bibitem{Vaswani2017attention}
Ashish Vaswani, Noam Shazeer, Niki Parmar, Jakob Uszkoreit, Llion Jones,
  Aidan~N. Gomez, Lukasz Kaiser, and Illia Polosukhin.
\newblock Attention is all you need.
\newblock In {\em Advances in Neural Information Processing Systems}, pages
  5998--6008, 2017.

\bibitem{Wang2017resatt}
Fei Wang, Mengqing Jiang, Chen Qian, Shuo Yang, Cheng Li, Honggang Zhang,
  Xiaogang Wang, and Xiaoou Tang.
\newblock Residual attention network for image classification.
\newblock In {\em {IEEE} Conference on Computer Vision and Pattern
  Recognition}, pages 6450--6458. {IEEE} Computer Society, 2017.

\bibitem{wang2019deep}
Yue Wang and Justin Solomon.
\newblock Deep closest point: Learning representations for point cloud
  registration.
\newblock In {\em {IEEE/CVF} International Conference on Computer Vision},
  pages 3522--3531. {IEEE}, 2019.

\bibitem{wang2019dynamic}
Yue Wang, Yongbin Sun, Ziwei Liu, Sanjay~E. Sarma, Michael~M. Bronstein, and
  Justin~M. Solomon.
\newblock Dynamic graph {CNN} for learning on point clouds.
\newblock {\em {ACM} Transactions on Graphics}, 38(5):146:1--146:12, 2019.

\bibitem{Wu2020visual}
Bichen Wu, Chenfeng Xu, Xiaoliang Dai, Alvin Wan, Peizhao Zhang, Masayoshi
  Tomizuka, Kurt Keutzer, and Peter Vajda.
\newblock Visual transformers: Token-based image representation and processing
  for computer vision.
\newblock {\em CoRR}, abs/2006.03677, 2020.

\bibitem{Wu2019pointconv}
Wenxuan Wu, Zhongang Qi, and Fuxin Li.
\newblock {PointConv}: Deep convolutional networks on 3d point clouds.
\newblock In {\em {IEEE/CVF} Conference on Computer Vision and Pattern
  Recognition}, pages 9621--9630, 2019.

\bibitem{Wu2015modelnet}
Zhirong Wu, Shuran Song, Aditya Khosla, Fisher Yu, Linguang Zhang, Xiaoou Tang,
  and Jianxiong Xiao.
\newblock 3d shapenets: {A} deep representation for volumetric shapes.
\newblock In {\em {IEEE} Conference on Computer Vision and Pattern Recognition,
  {CVPR} 2015, Boston, MA, USA, June 7-12, 2015}, pages 1912--1920. {IEEE}
  Computer Society, 2015.

\bibitem{Xie_2018_CVPR}
Saining Xie, Sainan Liu, Zeyu Chen, and Zhuowen Tu.
\newblock Attentional shapecontextnet for point cloud recognition.
\newblock In {\em Proceedings of the IEEE Conference on Computer Vision and
  Pattern Recognition (CVPR)}, June 2018.

\bibitem{yan2020pointasnl}
Xu Yan, Chaoda Zheng, Zhen Li, Sheng Wang, and Shuguang Cui.
\newblock {PointASNL}: Robust point clouds processing using nonlocal neural
  networks with adaptive sampling.
\newblock In {\em {IEEE/CVF} Conference on Computer Vision and Pattern
  Recognition}, pages 5588--5597. {IEEE}, 2020.

\bibitem{pan2018convolutional}
Yuqi Yang, Shilin Liu, Hao Pan, Yang Liu, and Xin Tong.
\newblock {PFCNN:} convolutional neural networks on 3d surfaces using parallel
  frames.
\newblock In {\em {IEEE/CVF} Conference on Computer Vision and Pattern
  Recognition}, pages 13575--13584. {IEEE}, 2020.

\bibitem{Yang2019xlnet}
Zhilin Yang, Zihang Dai, Yiming Yang, Jaime~G. Carbonell, Ruslan Salakhutdinov,
  and Quoc~V. Le.
\newblock Xlnet: Generalized autoregressive pretraining for language
  understanding.
\newblock In Hanna~M. Wallach, Hugo Larochelle, Alina Beygelzimer, Florence
  d'Alch{\'{e}}{-}Buc, Emily~B. Fox, and Roman Garnett, editors, {\em Advances
  in Neural Information Processing Systems}, pages 5754--5764, 2019.

\bibitem{yi2016shapenet}
Li Yi, Vladimir~G. Kim, Duygu Ceylan, I{-}Chao Shen, Mengyan Yan, Hao Su, Cewu
  Lu, Qixing Huang, Alla Sheffer, and Leonidas~J. Guibas.
\newblock A scalable active framework for region annotation in 3d shape
  collections.
\newblock {\em {ACM} Trans. Graph.}, 35(6):210:1--210:12, 2016.

\bibitem{Zhang2019sagan}
Han Zhang, Ian~J. Goodfellow, Dimitris~N. Metaxas, and Augustus Odena.
\newblock Self-attention generative adversarial networks.
\newblock In Kamalika Chaudhuri and Ruslan Salakhutdinov, editors, {\em
  International Conference on Machine Learning}, volume~97 of {\em Proceedings
  of Machine Learning Research}, pages 7354--7363. {PMLR}, 2019.

\bibitem{Zhao2019pointweb}
Hengshuang Zhao, Li Jiang, Chi{-}Wing Fu, and Jiaya Jia.
\newblock Pointweb: Enhancing local neighborhood features for point cloud
  processing.
\newblock In {\em {IEEE} Conference on Computer Vision and Pattern
  Recognition}, pages 5565--5573. Computer Vision Foundation / {IEEE}, 2019.

\end{thebibliography}
}

\end{document}